\documentclass{article}

\usepackage[final]{corl_2019} %
\usepackage{microtype}
\usepackage{graphicx}
\usepackage{mathtools}
\usepackage{siunitx}
\usepackage[font=small]{caption}
\usepackage{booktabs} %
\usepackage{amssymb}
\usepackage{amsmath,amsthm}
\usepackage{mathtools}
\usepackage{subcaption}
\usepackage{algorithm}
\usepackage{algpseudocode}
\usepackage[utf8]{inputenc}
\usepackage[english]{babel}
\usepackage{enumitem}
\newtheorem{theorem}{Theorem}[section]
\newtheorem{lemma}{Lemma}[section]
\newtheorem{corollary}{Corollary}[section]
\theoremstyle{definition}
\theoremstyle{assumption}
\newtheorem{definition}{Definition}[section]
\newtheorem{assumption}{Assumption}[section]
\usepackage{hyperref}
\usepackage{cleveref}

\newcommand{\todo}[1]{}
\renewcommand{\todo}[1]{{\color{red} TODO: {#1}}}

\DeclareMathOperator*{\argmin}{arg\,min}

\title{On-Policy Robot Imitation Learning from a Converging Supervisor}

\author{
  Ashwin Balakrishna*, Brijen Thananjeyan*, Jonathan Lee, Felix Li, Arsh Zahed, \\ \textbf{Joseph E. Gonzalez, Ken Goldberg}\\
  Department of Electrical Engineering and Computer Sciences\\
  University of California Berkeley 
  United States\\
  \texttt{ashwin\_balakrishna@berkeley.edu, bthananjeyan@berkeley.edu} \\\scriptsize{* equal contribution} \\
}
\begin{document}
\newcommand{\algname}{Follow the Improving Teacher (FIT)\xspace}
\newcommand{\justalgname}{Follow the Improving Teacher\xspace}
\newcommand{\algabbr}{FIT}
\newcommand\smallO{
  \mathchoice
    {{\scriptstyle\mathcal{O}}}%
    {{\scriptstyle\mathcal{O}}}%
    {{\scriptscriptstyle\mathcal{O}}}%
    {\scalebox{.6}{$\scriptscriptstyle\mathcal{O}$}}%
  }

\maketitle

\begin{abstract}
Existing on-policy imitation learning algorithms, such as DAgger, assume access to a fixed supervisor. However, there are many settings where the supervisor may evolve during policy learning, such as a human performing a novel task or an improving algorithmic controller. We formalize imitation learning from a ``converging supervisor" and provide sublinear static and dynamic regret guarantees against the best policy in hindsight with labels from the converged supervisor, even when labels during learning are only from intermediate supervisors. We then show that this framework is closely connected to a class of reinforcement learning (RL) algorithms known as dual policy iteration (DPI), which alternate between training a reactive learner with imitation learning and a model-based supervisor with data from the learner. Experiments suggest that when this framework is applied with the state-of-the-art deep model-based RL algorithm PETS as an improving supervisor, it outperforms deep RL baselines on continuous control tasks and provides up to an 80-fold speedup in policy evaluation.
\end{abstract}

\keywords{Imitation Learning, Online Learning, Reinforcement Learning}

\section{Introduction}

In robotics there is significant interest in using human or algorithmic supervisors to train policies via imitation learning \cite{one-shot-visual-IL, imitation-context, imitation-observe, imitation-virtual}. For example, a trained surgeon with experience teleoperating a surgical robot can provide successful demonstrations of surgical maneuvers~\cite{gao2014jhu}. Similarly, known dynamics models can be used by standard control techniques, such as model predictive control (MPC), to generate controls to optimize task reward \cite{PLATO, pan2017agile}. However, there are many cases in which the supervisor is not fixed, but is \textit{converging} to improved behavior over time, such as when a human is initially unfamiliar with a teleoperation interface or task or when the dynamics of the system are initially unknown and estimated with experience from the environment when training an algorithmic controller. Furthermore, these supervisors are often \textit{slow}, as humans can struggle to execute stable, high-frequency actions on a robot \cite{pan2017agile} and model-based control techniques, such as MPC, typically require computationally expensive stochastic optimization techniques to plan over complex dynamics models~\cite{handful-of-trials, NNDynamics, saved2019}. This motivates algorithms that can distill supervisors which are both \textit{converging} and \textit{slow} into policies that can be efficiently executed in practice. The idea of distilling improving algorithmic controllers into reactive policies has been explored in a class of reinforcement learning (RL) algorithms known as dual policy iteration (DPI)~\cite{anthony2017thinking,silver2017mastering, sun2018dual}, which alternate between optimizing a reactive learner with imitation learning and a model-based supervisor with data from the learner. However, past methods have mostly been applied in discrete settings \cite{anthony2017thinking, silver2017mastering} or make specific structural assumptions on the supervisor \cite{sun2018dual}.

This paper analyzes learning from a converging supervisor in the context of on-policy imitation learning. Prior analysis of on-policy imitation learning algorithms provide regret guarantees given a fixed supervisor \cite{DAgger, deeply-aggrevated, Lee2019ADR, convergence-value-agg}. We consider a converging sequence of supervisors and show that similar guarantees hold for the regret against the best policy in hindsight with labels from the converged supervisor, even when only intermediate supervisors provide labels during learning. Since the analysis makes no structural assumptions on the supervisor, this flexibility makes it possible to use any off-policy method as the supervisor in the presented framework, such as an RL algorithm or a human, provided that it converges to a good policy on the learner's distribution. We implement an instantiation of this framework with the deep MPC algorithm PETS \cite{handful-of-trials} as an improving supervisor and maintain the data efficiency of PETS while significantly reducing online computation time, accelerating both policy learning and evaluation.

The key contribution of this work is a new framework for on-policy imitation learning from a converging supervisor. We present a new notion of static and dynamic regret in this setting and provide sublinear regret guarantees by showing a reduction from this new notion of regret to the standard notion for the fixed supervisor setting. The dynamic regret result is particularly unintuitive, as it indicates that it is possible to do well on each round of learning compared to a learner with labels from the converged supervisor, even though labels are only provided by intermediate supervisors during learning. We then show that the presented framework relaxes assumptions on the supervisor in DPI and perform simulated continuous control experiments suggesting that when a PETS supervisor~\cite{handful-of-trials} is used, we can outperform other deep RL baselines while achieving up to an 80-fold speedup in policy evaluation. Experiments on a physical surgical robot yield up to an 20-fold reduction in query time and 53\% reduction in policy evaluation time after accounting for hardware constraints.
\section{Related Work}

On-policy imitation learning algorithms that directly learn reactive policies from a supervisor were popularized with DAgger~\cite{ross2010reduction}, which iteratively improves the learner by soliciting supervisor feedback on the learner's trajectory distribution. This yields significant performance gains over analogous off-policy methods \cite{Bagnell-2015-5921,pomerleau1989alvinn}. On-policy methods have been applied with both human \cite{laskey2017comparing} and algorithmic supervisors \cite{pan2017agile}, but with a fixed supervisor as the guiding policy. We propose a setting where the supervisor improves over time, which is common when learning from a human or when distilling a computationally expensive, iteratively improving controller into a policy that can be efficiently executed in practice. Recently, convergence results and guarantees on regret metrics such as dynamic regret have been shown for the fixed supervisor setting \cite{Lee2019ADR, convergence-value-agg, Cheng-COL}. We extend these results and present a static and dynamic analysis of on-policy imitation learning from a convergent sequence of supervisors. Recent work proposes using inverse RL to outperform an improving supervisor~\cite{pmlr-v97-jacq19a, TREX}. We instead study imitation learning in this context to use an evolving supervisor for policy learning.

Model-based planning has seen significant interest in RL due to the benefits of leveraging structure in settings such as games and robotic control \cite{anthony2017thinking,silver2017mastering, sun2018dual}. Deep model-based reinforcement learning (MBRL) has demonstrated 
superior data efficiency compared to model-free methods and state-of-the-art performance on a variety of continuous control tasks ~\cite{handful-of-trials, NNDynamics, saved2019}. However, these techniques are often too computationally expensive for high-frequency execution, significantly slowing down policy evaluation. To address the online burden of model-based algorithms, \citet{sun2018dual} define a novel class of algorithms, dual policy iteration (DPI), which alternate between optimizing a fast learner for policy evaluation using labels from a model-based supervisor and optimizing a slower model-based supervisor using trajectories from the learner. However, past work in DPI either involves planning in discrete state spaces \cite{anthony2017thinking, silver2017mastering}, or making specific assumptions on the structure of the model-based controller \cite{sun2018dual}. We discuss how the converging supervisor framework is connected to DPI, but enables a more flexible supervisor specification. We then provide a practical algorithm by using the deep MBRL algorithm PETS \cite{handful-of-trials} as an improving supervisor to achieve fast policy evaluation while maintaining the data efficiency of PETS.
\section{Converging Supervisor Framework and Preliminaries}
\label{sec:framework}

\subsection{On-Policy Imitation Learning}
\label{sec:OPIL}
We consider continuous control problems in a finite-horizon Markov decision process (MDP), which is defined by a tuple $(\mathcal{S}, \mathcal{A}, P(\cdot, \cdot), T, R(\cdot,\cdot))$ where $\mathcal{S}$ is the state space and $\mathcal{A}$ is the action space. The stochastic dynamics model $P$ maps a state $s$ and action $a$ to a probability distribution over states, $T$ is the task horizon, and $R$ is the reward function. A deterministic control policy $\pi$ maps an input state in $\mathcal{S}$ to an action in $\mathcal{A}$. The goal in RL is to learn a policy $\pi$ over the MDP which induces a trajectory distribution that maximizes the sum of rewards along the trajectory. In imitation learning, this objective is simplified by instead optimizing a surrogate loss function which measures the discrepancy between the actions chosen by learned parameterized policy $\pi_\theta$ and supervisor $\psi$.

Rather than directly optimizing $R$ from experience, on-policy imitation learning involves executing a policy in the environment and then soliciting feedback from a supervisor on the visited states. This is in contrast to off-policy methods, such as behavior cloning, in which policy learning is performed entirely on states from the supervisor's trajectory distribution. The surrogate loss of a policy $\pi_\theta$ along a trajectory is a supervised learning cost defined by the supervisor relabeling the trajectory's states with actions. The goal of on-policy imitation is to find the policy minimizing the corresponding surrogate risk on its own trajectory distribution. On-policy algorithms typically adhere to the following iterative procedure: (1) at iteration $i$, execute the current policy $\pi_{\theta_i}$ by deploying the learner in the MDP, observing states and actions as trajectories; (2) Receive labels for each state from the supervisor $\psi$; (3) Update $\pi_{\theta_i}$ according to the supervised learning loss to generate $\pi_{\theta_{i+1}}$.

On-policy imitation learning has often been viewed as an instance of online optimization or online learning \cite{DAgger, Lee2019ADR, convergence-value-agg}. Online optimization is posed as a game between an adversary, which generates a loss function $l_i$ at iteration $i$ and an algorithm, which plays a policy $\pi_{\theta_i}$ in an attempt to minimize the total incurred losses. After observing $l_i$, the algorithm updates its policy $\pi_{\theta_{i+1}}$ for the next iteration. In the context of imitation learning, the loss $l_i(\cdot)$ at iteration $i$ corresponds to the supervised learning loss function under the current policy. The loss function $\l_i(\cdot)$ can then be used to update the policy for the next iteration. The benefit of reducing on-policy imitation learning to online optimization is that well-studied analyses and regret metrics from online optimization can be readily applied to understand and improve imitation learning algorithms. Next, we outline a theoretical framework in which to study on-policy imitation learning with a converging supervisor.

\subsection{Converging Supervisor Framework (CSF)}

We begin by presenting a set of definitions for on-policy imitation learning with a converging supervisor in order to analyze the static regret (Section \ref{sec:static-regret}) and dynamic regret (Section \ref{sec:dynamic-regret}) that can be achieved in this setting. In this paper, we assume that policies $\pi_\theta$ are parameterized by a parameter $\theta$ from a convex compact set $\Theta \subset \mathbb R^d$ equipped with the $l_2$-norm, which we denote with $\lVert \cdot \rVert$ for simplicity for both vectors and operators.

\begin{definition}
\label{def-supervisor}
\textbf{Supervisor: } We can think of a converging supervisor as a sequence of supervisors (labelers), $(\psi_i)_{i=1}^{\infty}$, where $\psi_i$ defines a deterministic controller $\psi_i: \mathcal{S} \rightarrow \mathcal{A}$. Supervisor $\psi_i$ provides labels for imitation learning policy updates at iteration $i$. 
\end{definition}
\begin{definition}
\label{def-learner}
\textbf{Learner: }The learner is represented at iteration $i$ by a parameterized policy $\pi_{\theta_i}: \mathcal{S} \rightarrow \mathcal{A}$ where $\pi_{\theta_i}$ is differentiable function in the policy parameter $\theta_i \in \Theta$.
\end{definition}
We denote the state and action at timestep $t$ in the trajectory $\tau$ sampled at iteration $i$ by the learner with $s_t^i$ and $a_t^i$ respectively.
\begin{definition}
\label{def-losses}
\textbf{Losses: }We consider losses at each round $i$ of the form: $l_{i}(\pi_{\theta}, \psi_i) = \mathbb{E}_{\tau \sim p(\tau|\theta_i)} \left[ \frac{1}{T} \sum_{t = 1}^{T} \lVert \pi_\theta(s^i_t) - \psi_i(s^i_t)\rVert^2 \right]$ where $p(\tau|\theta_i)$ defines the distribution of trajectories generated by $\pi_{\theta_i}$. Gradients of $l_i$ with respect to $\theta$ are defined as $\nabla_\theta l_i(\pi_{\theta_i}, \psi_i) = \nabla_\theta l_i(\pi_{\theta}, \psi_i)\bigr\rvert_{\theta=\theta_i}$.
\end{definition}

For analysis of the converging supervisor setting, we adopt the following standard assumptions. The assumptions in this section and the loss formulation are consistent with those in~\citet{hazan2016introduction} and~\citet{DAgger} for analysis of online optimization and imitation learning algorithms. The loss incurred by the agent is the population risk of the policy, and extension to empirical risk can be derived via standard concentration inequalities as in~\citet{DAgger}.

\begin{assumption}
\label{assum-strongly-convex}
\textbf{Strongly convex losses:} $\forall \theta_i \in \Theta,\ l_{i}(\pi_{\theta}, \psi)$ is strongly convex with respect to $\theta$ with parameter $\alpha \in \mathbb{R}^+$. Precisely, we assume that
\begin{align*} 
\begin{split}
&l_{i}(\pi_{\theta_2}, \psi) \geq l_{i}(\pi_{\theta_1}, \psi) + \nabla_{\theta} l_{i}(\pi_{\theta_1}, \psi)^T (\theta_2 - \theta_1) + \frac{\alpha}{2} \lVert \theta_2 - \theta_1 \rVert^2 \ \ \ \forall \ \theta_1, \theta_2 \in \Theta
\end{split}
\end{align*}
\end{assumption}
The expectation over $p(\tau|\theta_i)$ in Assumption \ref{assum-strongly-convex} preserves strong convexity of the squared loss for an individual sample, which is assumed to be convex in $\theta$.

\begin{assumption}
\label{assum-bounded-op-norm}
\textbf{Bounded operator norm of policy Jacobian: }$\lVert \nabla_{\theta}\pi_{\theta_i}(s)\rVert \leq G$ $\ \ \forall s \in \mathcal{S}$, $\ \ \forall \ \theta, \theta_i \in \Theta$ where $G \in \mathbb{R}^+$.
\end{assumption}

\begin{assumption}
\label{assum-bounded-spaces}
\textbf{Bounded action space:} The action space $\mathcal{A}$ has diameter $\delta$. Equivalently stated: $\delta = \sup_{a_1,a_2\in\mathcal{A}}\|a_1 - a_2\|$.
\end{assumption}

\section{Regret Analysis}

We analyze the performance of well-known algorithms in on-policy imitation learning and online optimization under the converging supervisor framework. In this setting, we emphasize that the goal is to achieve low loss $l_i(\pi_{\theta_i}, \psi_N)$ with respect to labels from the last observed supervisor $\psi_N$. We achieve these results through regret analysis via reduction of on-policy imitation learning to online optimization, where regret is a standard notion for measuring the performance of algorithms. We consider two forms: static and dynamic regret \cite{zinkevich2003online}, both of which have been utilized in previous on-policy imitation learning analyses \cite{DAgger,Lee2019ADR}. In this paper, regret is defined with respect to the expected losses under the trajectory distribution induced by the realized sequence of policies $(\pi_{\theta_i})_{i=1}^{N}$. Standard concentration inequalities can be used for finite sample analysis as in~\citet{DAgger}.

Using static regret, we can show a loose upper bound on average performance with respect to the last observed supervisor with minimal assumptions, similar to \cite{DAgger}. Using dynamic regret, we can tighten this upper bound, showing that $\theta_i$ is optimal in expectation on its own distribution with respect to $\psi_N$ for certain algorithms, similar to \cite{Lee2019ADR,Cheng-COL}; however, to achieve this stronger result, we require an additional continuity assumption on the dynamics of the system, which was shown to be necessary by \citet{convergence-value-agg}. To harness regret analysis in imitation learning, we seek to show that algorithms achieve \textit{sublinear regret} (whether static or dynamic), denoted by $\smallO(N)$ where $N$ is the number of iterations. That is, the regret should grow at a slower rate than linear in the number of iterations. While existing algorithms can achieve sublinear regret in the fixed supervisor setting, we analyze regret with respect to the last observed supervisor $\psi_N$, even though the learner is only provided labels from the intermediate ones during learning. See supplementary material for all proofs.

\subsection{Static Regret}
\label{sec:static-regret}
Here we show that as long as the supervisor labels are Cauchy, i.e. if $\forall s \in \mathcal{S},\ \forall N > i, \lVert \psi_i(s) - \psi_N(s) \rVert \leq f_i$ where $\lim_{i \rightarrow \infty} f_i = 0$, it is possible to achieve sublinear static regret with respect to the best policy in hindsight with labels from  $\psi_N$ for the whole dataset. This is a more difficult metric than is typically considered in regret analysis for on-policy imitation learning since labels are provided by the converging supervisor $\psi_i$ at iteration $i$, but regret is evaluated with respect to the best policy given labels from $\psi_N$. Past work has shown that it is possible to obtain sublinear static regret in the fixed supervisor setting under strongly convex losses for standard on-policy imitation learning algorithms such as online gradient descent \cite{hazan2016introduction} and DAgger \cite{DAgger}; we extend this and show that the additional asymptotic regret in the converging supervisor setting depends only on the convergence rate of the supervisor. The standard notion of static regret is given in Definition \autoref{def-static-regret-standard}.

\begin{definition} \label{def-static-regret-standard}
The static regret with respect to the sequence of supervisors $(\psi_i)_{i=1}^{N}$ is given by the difference in the performance of policy $\pi_{\theta_i}$ and that of the best policy in hindsight under the average trajectory distribution induced by the incurred losses with labels from current supervisor $\psi_i$.
\begin{align*}
&\mathrm{Regret}_N^S ( (\psi_i)_{i=1}^{N}) = \sum_{i = 1}^{N} l_i(\pi_{\theta_i}, \psi_i) - \sum_{i = 1}^{N} l_i(\pi_{\theta^{\ast}}, \psi_i) \text{ where }\theta^\ast = \argmin_{\theta \in \Theta} \sum_{i = 1}^{N} l_i(\pi_{\theta}, \psi_i)
\end{align*}
\end{definition}

However, we instead analyze the more difficult regret metric presented in Definition \autoref{def-static-regret} below.

\begin{definition}
\label{def-static-regret}
The static regret with respect to the supervisor $\psi_N$ is given by the difference in the performance of policy $\pi_{\theta_i}$ and that of the best policy in hindsight under the average trajectory distribution induced by the incurred losses with labels from the last observed supervisor $\psi_N$.
\begin{align*}
&\mathrm{Regret}_N^S (\psi_N) = \sum_{i = 1}^{N} l_i(\pi_{\theta_i}, \psi_N) - \sum_{i = 1}^{N} l_i(\pi_{\theta^{\star}}, \psi_N) \text{ where }\theta^\star = \argmin_{\theta \in \Theta} \sum_{i = 1}^{N} l_i(\pi_{\theta}, \psi_N)
\end{align*}
\end{definition}

\begin{theorem}
\label{thm-static-regret-bound}
$\mathrm{Regret}_N^S (\psi_N)$ can be bounded above as follows:
\begin{align*} 
\begin{split}
\mathrm{Regret}_N^S (\psi_N) &\leq  \mathrm{Regret}_{N}^{S}((\psi_i)_{i=1}^N)
+ 4\delta \sum_{i=1}^{N} \mathbb{E}_{\tau \sim p(\tau|\theta_i)}\left[ \frac{1}{T} \sum_{t=1}^{T}\|\psi_N(s^i_t) - \psi_i(s^i_t)\| \right]
\end{split}
\end{align*}
\end{theorem}

\Cref{thm-static-regret-bound} essentially states that the expected static regret in the converging supervisor setting can be decomposed into two terms: one that is the standard notion of static regret, and an additional term that scales with the rate at which the supervisor changes. Thus, as long as there exists an algorithm to achieve sublinear static regret on the standard problem, the only additional regret comes from the evolution of the supervisor. Prior work has shown that algorithms such as online gradient descent \cite{hazan2016introduction} and DAgger \cite{DAgger} achieve sublinear static regret under strongly convex losses. Given this reduction, we see that these algorithms can also be used to achieve sublinear static regret in the converging supervisor setup if the extra term is sublinear. \Cref{lemma-static-regret-rate} identifies when this is the case.

\begin{corollary}
\label{lemma-static-regret-rate}
If $\ \forall s \in \mathcal{S},\ \forall N > i, \lVert \psi_i(s) - \psi_N(s)\rVert \leq f_i \text{ where }\lim_{i \rightarrow \infty} f_i = 0$, then $\mathrm{Regret}_N^S (\psi_N)$ can be decomposed as follows:
\begin{align*}
&\mathrm{Regret}_{N}^{S}(\psi_N) = \mathrm{Regret}_{N}^{S}((\psi_i)_{i=1}^N)  + \smallO(N)  
\end{align*}
\end{corollary}
\subsection{Dynamic Regret}
\label{sec:dynamic-regret}
Although the static regret analysis provides a bound on the average loss, the quality of that bound depends on the term $\min_{\theta} \sum_{i = 1}^N l_i(\pi_\theta,\psi_N)$, which in practice is often very large due to approximation error between the policy class and the actual supervisor. Furthermore, it has been shown that despite sublinear static regret, policy learning may be unstable under certain dynamics \cite{convergence-value-agg, laskey2017comparing}. Recent analyses have turned to dynamic regret \cite{Lee2019ADR, convergence-value-agg}, which measures the sub-optimality of a policy on its own distribution: $l_i(\pi_{\theta_i}, \psi_N) - \min_\theta l_i(\pi_\theta, \psi_N)$. Thus, low dynamic regret shows that a policy is on average performing optimally on its own distribution. This framework also helps determine if policy learning will be stable or if convergence is possible \cite{Lee2019ADR}. However, these notions require understanding the sensitivity of the MDP to changes in the policy. We quantify this with an additional Lipschitz assumption on the trajectory distributions induced by the policy as in \cite{Lee2019ADR, convergence-value-agg, Cheng-COL}. We show that even in the converging supervisor setting, it is possible to achieve sublinear dynamic regret given this additional assumption and a converging supervisor by reducing the problem to a predictable online learning problem \cite{Cheng-COL}. Note that this yields the surprising result that it is possible to do well on each round even against a dynamic comparator which has labels from the last observed supervisor. The standard notion of dynamic regret is given in \Cref{def-dynamic-regret-standard} below.

\begin{definition} \label{def-dynamic-regret-standard}
The dynamic regret with respect to the sequence of supervisors $(\psi_i)_{i=1}^{N}$ is given by the difference in the performance of policy $\pi_{\theta_i}$ and that of the best policy under the current round's loss, which compares the performance of current policy $\pi_{\theta_i}$ and current supervisor $\psi_i$.
\begin{align*}
&\mathrm{Regret}_{N}^{D}((\psi_i)_{i=1}^N) = \sum_{i = 1}^{N} l_{i}(\pi_{\theta_i}, \psi_i) - \sum_{i = 1}^{N} l_{i}(\pi_{\theta^{\ast}_i}, \psi_i) \text{ where }\theta^\ast_i = \argmin_{\theta \in \Theta} l_{i}(\pi_{\theta}, \psi_i)    
\end{align*}
\end{definition}

However, similar to the static regret analysis in Section \ref{sec:static-regret}, we seek to analyze the dynamic regret with respect to labels from the last observed supervisor $\psi_N$, which is defined as follows.
\begin{definition} \label{def-dynamic-regret}
The dynamic regret with respect to supervisor $\psi_N$ is given by the difference in the performance of policy $\pi_{\theta_i}$ and that of the best policy under the current round's loss, which compares the performance of current policy $\pi_{\theta_i}$ and last observed supervisor $\psi_N$.
\begin{align*}
&\mathrm{Regret}_{N}^{D}(\psi_N) = \sum_{i = 1}^{N} l_{i}(\pi_{\theta_i}, \psi_N) - \sum_{i = 1}^{N} l_{i}(\pi_{\theta^{\star}_i}, \psi_N) \text{ where }\theta^\star_i = \argmin_{\theta \in \Theta} l_{i}(\pi_{\theta}, \psi_N)    
\end{align*}
\end{definition}

We first show that there is a reduction from $\mathrm{Regret}_{N}^{D}(\psi_N)$ to $\mathrm{Regret}_{N}^{D}((\psi_i)_{i=1}^N)$.

\begin{lemma}
\label{lemma-dynamic-regret-reduction}
$\mathrm{Regret}_{N}^{D}(\psi_N)$ can be bounded above as follows:
\begin{align*}
\mathrm{Regret}_{N}^{D}(\psi_N) &\leq \mathrm{Regret}_{N}^{D}((\psi_i)_{i=1}^N) + 4\delta \sum_{i=1}^{N} \mathbb{E}_{\tau \sim p(\tau|\theta_i)}\left[ \frac{1}{T} \sum_{t=1}^{T}\|\psi_N(s^i_t) - \psi_i(s^i_t)\| \right]
\end{align*}
\end{lemma}

Given the notion of supervisor convergence discussed in \Cref{lemma-static-regret-rate}, \Cref{lemma-sublinear-standard-dynamic-regret} shows that if we can achieve sublinear $\mathrm{Regret}_{N}^{D}((\psi_i)_{i=1}^N)$, we can also achieve sublinear $\mathrm{Regret}_{N}^{D}(\psi_N)$.

\begin{corollary}
\label{lemma-sublinear-standard-dynamic-regret}

If $\ \forall s \in \mathcal{S},\ \forall N > i, \lVert \psi_i(s) - \psi_N(s)\rVert \leq f_i \text{ where }\lim_{i \rightarrow \infty} f_i = 0$, then $\mathrm{Regret}_{N}^{D}(\psi_N)$ can be decomposed as follows:
\begin{align*}
&\mathrm{Regret}_{N}^{D}(\psi_N) = \mathrm{Regret}_{N}^{D}((\psi_i)_{i=1}^N)  + \smallO(N)  
\end{align*}
\end{corollary}

It is well known that $\mathrm{Regret}_{N}^{D}((\psi_i)_{i=1}^N)$ cannot be sublinear in general \cite{Lee2019ADR}. However, as in \cite{Lee2019ADR, convergence-value-agg}, we can obtain conditions for sublinear regret by leveraging the structure in the imitation learning problem with a Lipschitz continuity condition on the trajectory distribution. Let $d_{TV}(p, q) = \frac{1}{2}\int |p - q| d\tau$ denote the total variation distance between two trajectory distributions $p$ and $q$.
\begin{assumption}
\label{assum-lipschitz-traj}
There exists $\eta \geq 0$ such that the following holds on the trajectory distributions induced by policies parameterized by $\theta_1$ and $\theta_2$:
\begin{align*}
    d_{TV}(p(\tau | \theta_1), p(\tau | \theta_2)) \leq \eta \| \theta_1 - \theta_2\|/2
\end{align*}
\end{assumption}

A similar assumption is made by popular RL algorithms~\cite{schulman2015trust,ME-TRPO}, and \Cref{lemma-dynamic-regret-standard} shows that with it, sublinear $\mathrm{Regret}_{N}^{D}((\psi_i)_{i=1}^N)$ can be achieved using results from predictable online learning \cite{Cheng-COL}.

\begin{lemma}
\label{lemma-dynamic-regret-standard}
If Assumption~\ref{assum-lipschitz-traj} holds and $\alpha > 4G\eta\sup_{a\in\mathcal{A}}\|a\|$, then there exists an algorithm where $\mathrm{Regret}_{N}^{D}((\psi_i)_{i=1}^N) = \smallO(N)$. If the diameter of the parameter space is bounded, the greedy algorithm, which plays $\theta_{i+1} = \argmin_{\theta \in \Theta} l_{i}(\pi_{\theta}, \psi_N)$, achieves sublinear $\mathrm{Regret}_{N}^{D}((\psi_i)_{i=1}^N)$. Furthermore, if the losses are $\gamma$-smooth in $\theta$ and $\frac{4G\eta\sup_{a\in\mathcal{A}}\|a\|}{\alpha} > \frac{\alpha}{2\gamma}$, then online gradient descent achieves sublinear $\mathrm{Regret}_{N}^{D}((\psi_i)_{i=1}^N)$.
\end{lemma}

Finally, we combine the results of \Cref{lemma-sublinear-standard-dynamic-regret} and \Cref{lemma-dynamic-regret-standard} to conclude that since we can achieve sublinear $\mathrm{Regret}_{N}^{D}((\psi_i)_{i=1}^N)$ and have found a reduction from $\mathrm{Regret}_{N}^{D}(\psi_N)$ to $\mathrm{Regret}_{N}^{D}((\psi_i)_{i=1}^N)$, we can also achieve sublinear dynamic regret in the converging supervisor setting. 

\begin{theorem} 
\label{thm-dynamic-regret-bound}
If $\ \forall s \in \mathcal{S},\ \forall N > i, \lVert \psi_i(s) - \psi_N(s)\rVert \leq f_i \text{ where }\lim_{i \rightarrow \infty} f_i = 0$ and under the assumptions in Lemma~\ref{lemma-dynamic-regret-standard}, there exists an algorithm where $\mathrm{Regret}_{N}^{D}(\psi_N) = \smallO(N)$. If the diameter of the parameter space is bounded, the greedy algorithm that plays $\theta_{i+1} = \argmin_{\theta \in \Theta} l_{i}(\pi_{\theta}, \psi_N)$ achieves sublinear $\mathrm{Regret}_{N}^{D}(\psi_N)$. Furthermore, if the losses are $\gamma$-smooth in $\theta$ and $\frac{4G\eta\sup_{a\in\mathcal{A}}\|a\|}{\alpha} > \frac{\alpha}{2\gamma}$, then online gradient descent achieves sublinear $\mathrm{Regret}_{N}^{D}(\psi_N)$.
\end{theorem}

\section{Converging Supervisors for Deep Continuous Control}
\label{sec:alg}
\citet{sun2018dual} apply DPI to continuous control tasks, but assume that both the learner and supervisor are of the same policy class and from a class of distributions for which computing the KL-divergence is computationally tractable. These constraints on supervisor structure limit model capacity compared to state-of-the-art deep RL algorithms. In contrast, we do not constrain the structure of the supervisor, making it possible to use any converging, improving supervisor (algorithmic or human) with no additional engineering effort. Note that while all provided guarantees only require that the supervisor \textit{converges}, we implicitly assume that the supervisor labels actually \textit{improve} with respect to the MDP reward function, $R$, when trained with data on the learner's distribution for the learner to achieve good task performance. This assumption is validated by the experimental results in this paper and those in prior work~\cite{anthony2017thinking,silver2017mastering}. One strategy to encourage the supervisor to improve on the learner's distribution is to add noise to the learner policy to increase the variety of the experience used by the supervisor to learn information such as system dynamics. However, this was not necessary for the environments considered in this paper, and we defer further study in this direction to future work.

We utilize the converging supervisor framework (CSF) to motivate an algorithm that uses the state-of-the-art deep MBRL algorithm, PETS, as an improving supervisor. Note that while for analysis we assume a deterministic supervisor, PETS produces stochastic supervision for the agent. We observe that this does not detrimentally impact performance of the policy in practice. PETS was chosen since it has demonstrated superior data efficiency compared to other deep RL algorithms \cite{handful-of-trials}. We collect policy rollouts from a model-free learner policy and refit the policy on each episode using DAgger \cite{DAgger} with supervision from PETS, which maintains a trained dynamics model based on the transitions collected by the learner. Supervision is generated via MPC by using the cross entropy method to plan over the learned dynamics for each state in the learner's rollout, but is collected after the rollout has completed rather than at each timestep of every policy rollout to reduce online computation time.
\section{Experiments}
\label{sec:exps}
The method presented in Section~\ref{sec:alg} uses the Converging Supervisor Framework (CSF) to train a learner policy to imitate a PETS supervisor trained on the learner's distribution. We expect the CSF learner to be less data efficient than PETS, but have significantly faster policy evaluation time. To evaluate this hypothesis, we measure the gap in data efficiency between the learner on its own distribution (CSF learner), the supervisor on the learner's distribution (CSF supervisor) and the supervisor on its own distribution (PETS).
Returns for the CSF learner and CSF supervisor are computed by rolling out the model-free learner policy and model-based controller after each training episode. Because the CSF supervisor is trained with off-policy data from the learner, the difference between the performance of the CSF learner and CSF supervisor measures how effectively the CSF learner is able to track the CSF supervisor's performance. The difference in performance between the CSF supervisor and PETS measures how important on-policy data is for PETS to generate good labels. All runs are repeated 3 times to control for stochasticity in training; see supplementary material for further experimental details. The DPI algorithm in \citet{sun2018dual} did not perform well on the presented environments, so we do not report a comparison to it.
However, we compare against the following set of 3 state-of-the-art model-free and model-based RL baselines and demonstrate that the CSF learner maintains the data efficiency of PETS while reducing online computation time significantly by only collecting policy rollouts from the fast model-free learner instead of from the PETS supervisor.
\begin{enumerate}[topsep=0pt,
noitemsep]
    \item \textbf{Soft Actor Critic (SAC)}: State-of-the-art maximum entropy model-free RL algorithm~\cite{SAC}.
    \item \textbf{Twin Delayed Deep Deterministic policy gradients (TD3)}: State-of-the-art model-free RL algorithm \cite{TD3} which uses target networks and delayed policy updates to improve DDPG~\cite{DDPG}, a popular actor critic algorithm.
    \item \textbf{Model-Ensemble Trust Region Policy Optimization (ME-TRPO)}: State-of-the-art model-free, model-based RL hybrid algorithm using a set of learned dynamics models to update a closed-loop policy offline with model-free RL~\cite{ME-TRPO}.
\end{enumerate}

\subsection{Simulation Experiments}
\begin{figure}
\centering
\begin{subfigure}[t]{.49\textwidth}
  \centering
  \includegraphics[width=\linewidth]{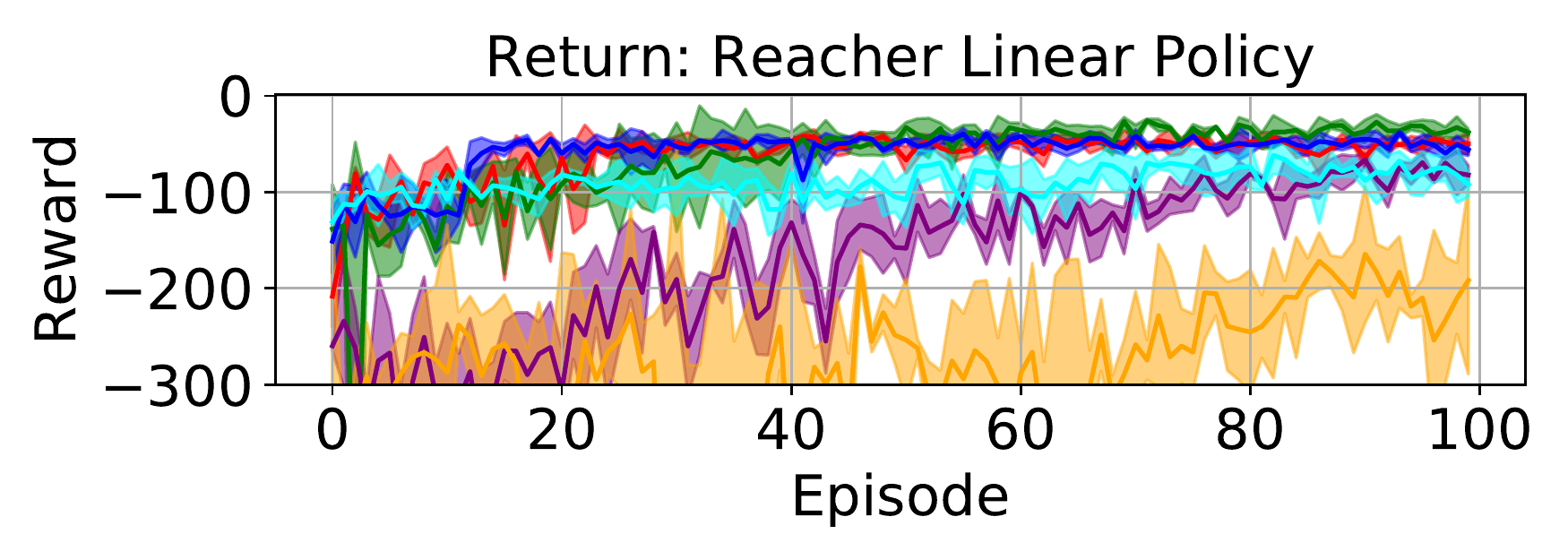}
\end{subfigure}
\hfill
\begin{subfigure}[t]{.49\textwidth}
  \centering
  \includegraphics[width=\linewidth]{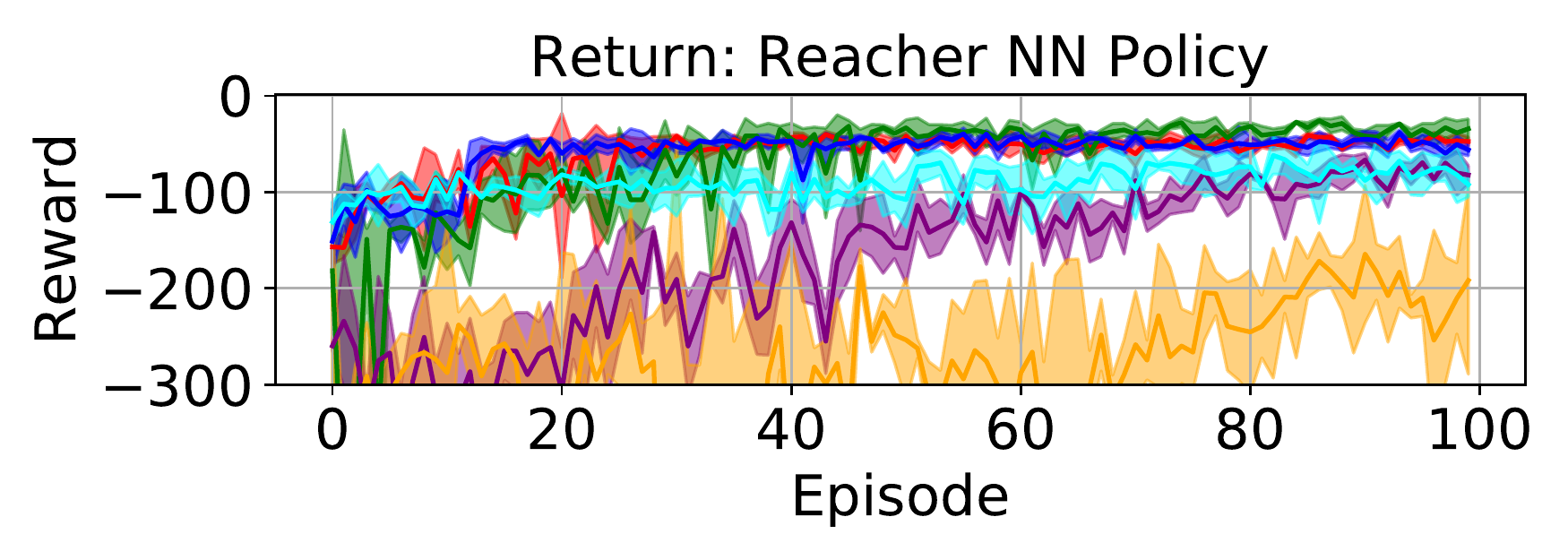}
\end{subfigure}
\vfill
\begin{subfigure}[t]{.49\textwidth}
  \centering
  \includegraphics[width=\linewidth]{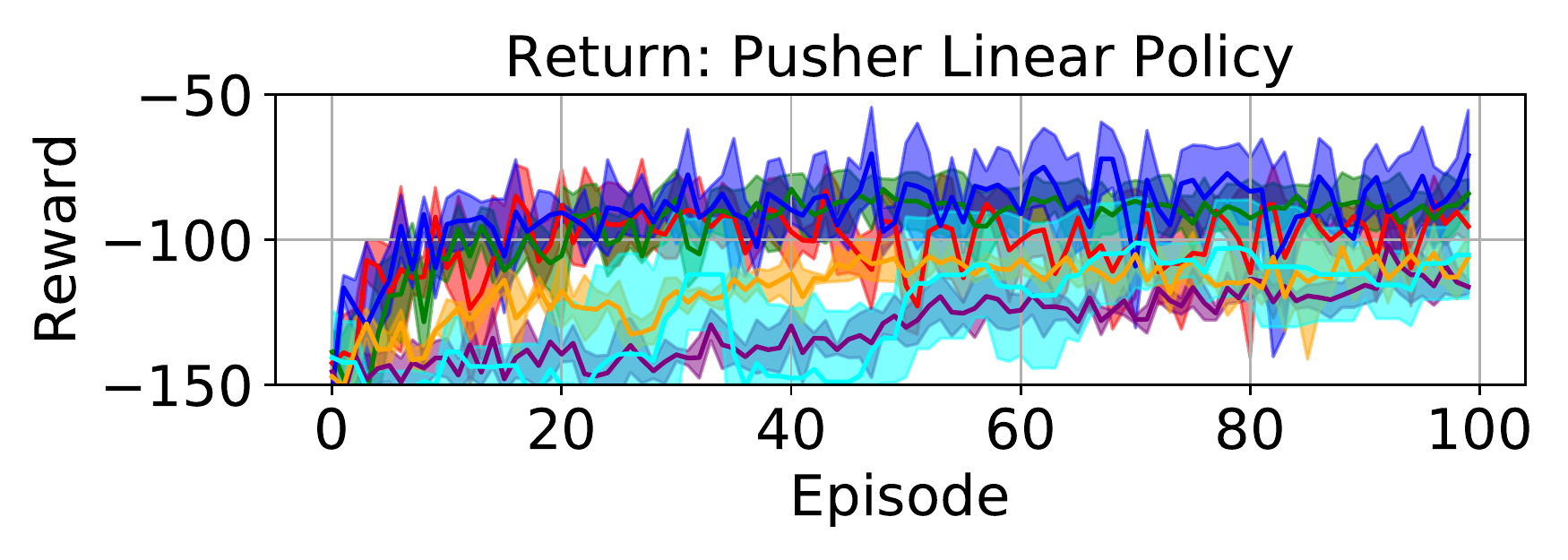}
\end{subfigure}
\hfill
\begin{subfigure}[t]{.49\textwidth}
  \centering
  \includegraphics[width=\linewidth]{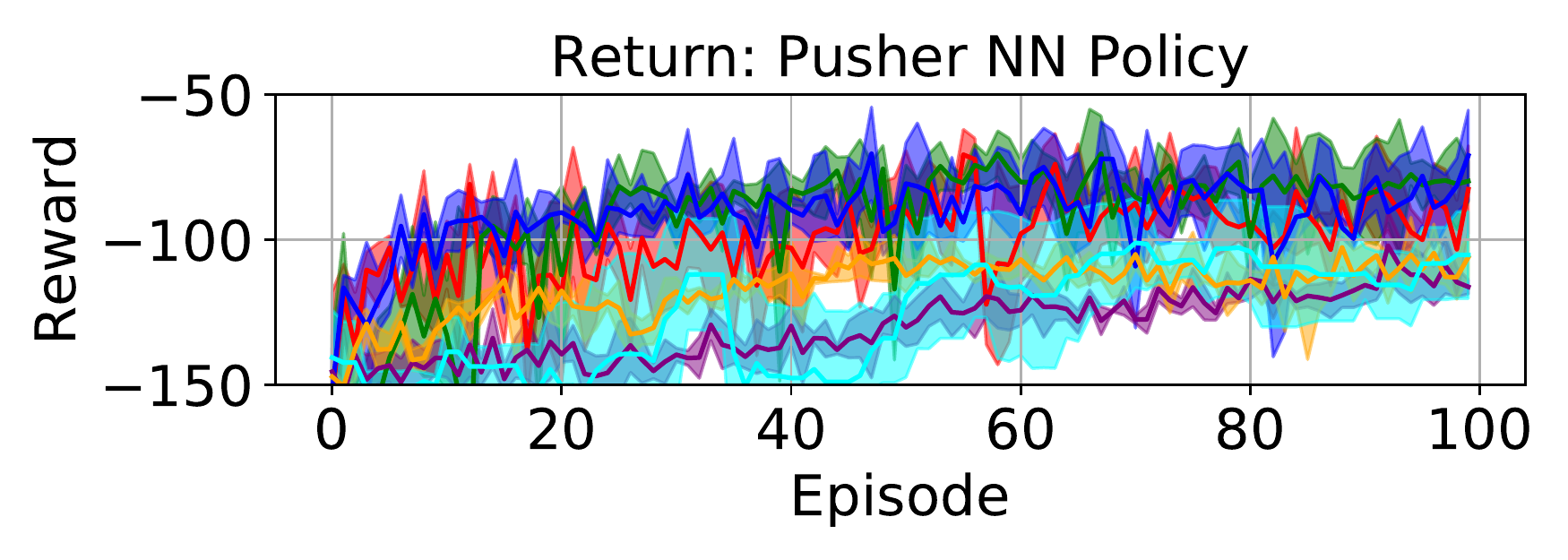}
\end{subfigure}
\vfill
\begin{subfigure}[t]{\textwidth}
  \centering
  \includegraphics[width=0.9\linewidth]{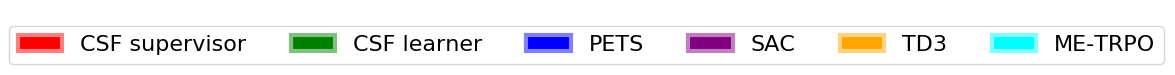}
\end{subfigure}

\caption{\textbf{Simulation experiments: } Training curves for the CSF learner, CSF supervisor, PETS, and baselines for the MuJoCo Reacher (top) and Pusher (bottom) tasks for a linear (left) and neural network (NN) policy (right). The linear policy is trained via ridge-regression with regularization parameter $\alpha = 1$, satisfying the strongly-convex loss assumption in Section~\ref{sec:framework}. To test more complex policy representations, we repeat experiments with a neural network (NN) learner with 2 hidden layers with 20 hidden units each. The CSF learner successfully tracks the CSF supervisor on both domains, performs well compared to PETS, and outperforms other baselines with both policy representations. The CSF learner is slightly less data efficient than PETS, but policy evaluation is up to 80x faster than PETS. SAC, TD3, and ME-TRPO use a neural network policy/dynamics class.
}\label{sim-figure}
\end{figure}

We consider the PR2 Reacher and Pusher continuous control MuJoCo domains from \citet{handful-of-trials} (Figure \ref{sim-figure}) since these are standard benchmarks on which PETS attains good performance. For both tasks, the CSF learner outperforms other state-of-the-art deep RL algorithms, demonstrating that the CSF learner enables fast policy evaluation while maintaining data efficient learning. The CSF learner closely matches the performance of both the CSF supervisor and PETS, indicating that the CSF learner has similar data efficiency as PETS. Results using a neural network CSF learner suggest that losses strongly-convex in $\theta$ may not be necessary in practice.

This result is promising because if the model-free learner policy is able to achieve similar performance to the supervisor on its own distribution, we can simultaneously achieve the data efficiency benefits of MBRL and the low online computation time of model-free methods. To quantify this speedup, we present timing results in Table \ref{timing-table}, which demonstrate that a significant speedup (up to 80x in this case) in policy evaluation is possible. Note that although we still need to evaluate the model-based controller on each state visited by the learner to generate labels, since this only needs to be done offline, this can be parallelized to reduce offline computation time as well.

\subsection{Physical Robot Experiments}

\begin{figure}
\centering
\begin{subfigure}[t]{.49\textwidth}
  \centering
  \includegraphics[width=\linewidth]{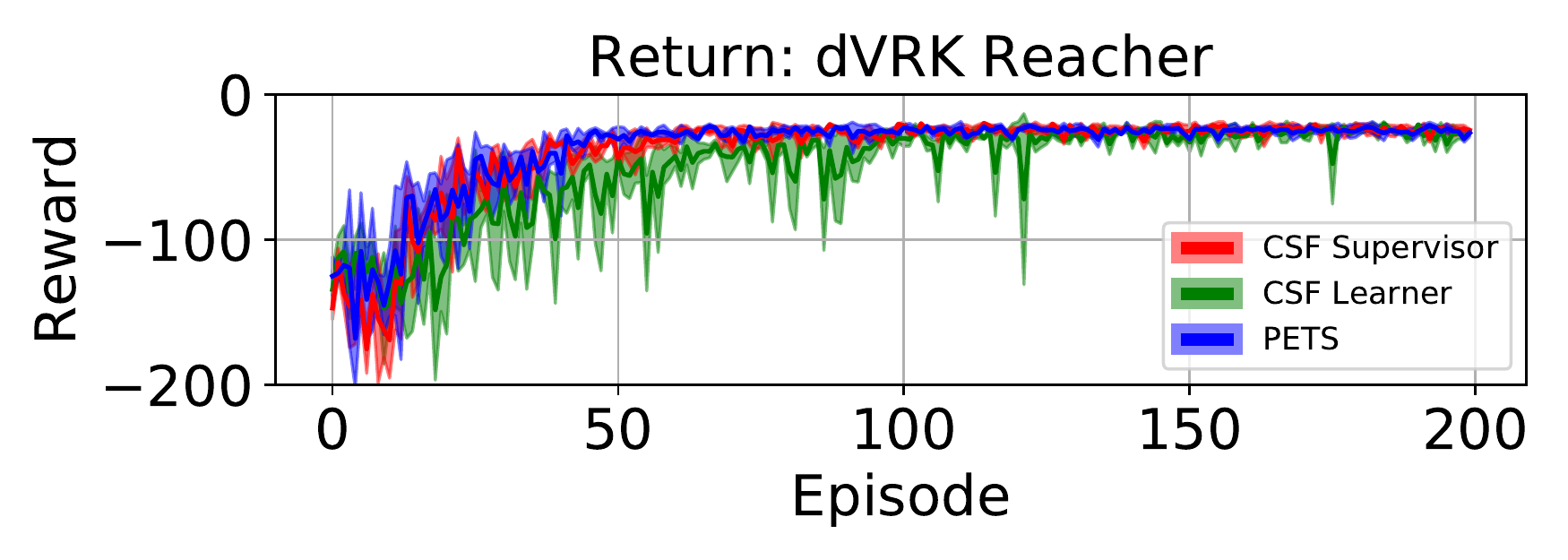}
\end{subfigure}
\hfill
\begin{subfigure}[t]{.49\textwidth}
  \centering
  \includegraphics[width=\linewidth]{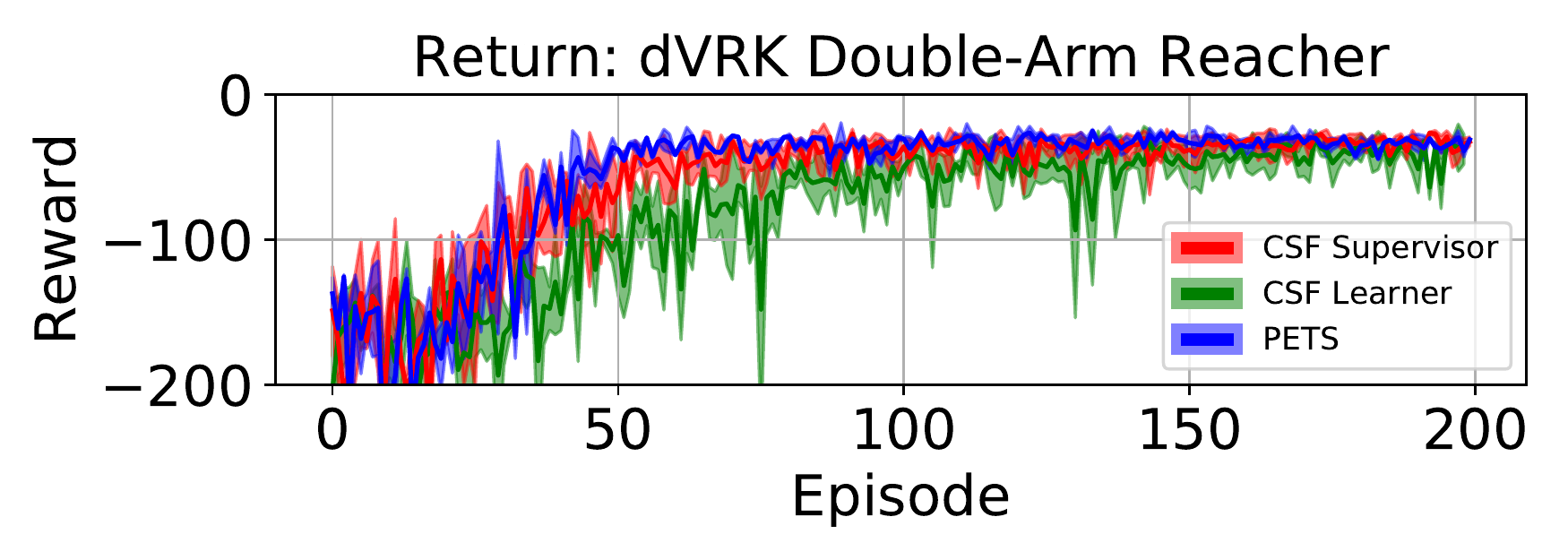}
\end{subfigure}
\caption{\textbf{Physical experiments: } Training curves for the CSF learner, CSF supervisor and PETS on the da Vinci surgical robot with a neural network policy. The CSF learner is able to track the CSF supervisor and PETS effectively and can be queried up to 20x faster than PETS. However, due to control frequency limitations on this system, the CSF learner has a policy evaluation time that is only 1.52 and 1.46 times faster than PETS for the single and double-arm tasks respectively. The performance gap between the CSF learner and the supervisor takes longer to diminish for the harder double-arm task.}
\label{real-figure}
\end{figure}

We also test CSF with a neural network policy on a physical da Vinci Surgical Robot (dVRK)~\cite{kazanzides-chen-etal-icra-2014} to evaluate its performance on multi-goal tasks where the end effector must be controlled to desired positions in the workspace. We evaluate the CSF learner/supervisor and PETS on the physical robot for both single and double arm versions of this task, and find that the CSF learner is able to track the PETS supervisor effectively (Figure~\ref{real-figure}) and provide up to a 22x speedup in policy query time (Table \ref{timing-table}). We expect the CSF learner to demonstrate significantly greater speedups relative to standard deep MBRL for higher dimensional tasks and for systems where higher-frequency commands are possible.

\begin{table}
\caption {\textbf{Policy evaluation and query times:} We report policy evaluation times in seconds over 100 episodes for the CSF learner and PETS (format: mean $\pm$ standard deviation). Furthermore, for physical experiments, we also report the total time taken to query the learner and PETS over an episode, since this difference in query times indicates the true speedup that CSF can enable (format: (total query time, policy evaluation time)). Policy evaluation and query times are nearly identical for simulation experiments. We see that the CSF learner is 20-80 times faster to query than PETS across all tasks. Results are reported on a desktop running Ubuntu 16.04 with a 3.60 GHz Intel Core i7-6850K and a NVIDIA GeForce GTX 1080. We use the NN policy for all timing results.}
\label{timing-table}
\begin{center}
\resizebox{\columnwidth}{!}{%
\begin{tabular}{ccccc}
\toprule
 & PR2 Reacher (Sim) & PR2 Pusher (Sim) & dVRK Reacher & dVRK Double-Arm Reacher\\
\midrule
CSF Learner & $\mathbf{0.29 \pm 0.01}$ & $\mathbf{1.13 \pm 0.66}$ & $\mathbf{(0.036\pm 0.009}, \mathbf{5.54 \pm 0.67})$ & $(\mathbf{0.038} \pm \mathbf{0.007}, \mathbf{8.87 \pm 1.12})$\\ 
PETS & $24.77 \pm 0.08$ & $57.77 \pm 17.12$ & $(0.78 \pm 0.02, 8.43 \pm 1.07)$ & $(0.88 \pm 0.07, 12.97 \pm 0.77)$\\ 
\bottomrule
\end{tabular}}
\end{center}
\vskip -0.3in
\end{table}
\section{Conclusion}
We formally introduce the converging supervisor framework for on-policy imitation learning and show that under standard assumptions, we can achieve sublinear static and dynamic regret against the best policy in hindsight with labels from the last observed supervisor, even when labels are only provided by the converging supervisor during learning. We then show a connection between the converging supervisor framework and DPI, and use this to present an algorithm to accelerate policy evaluation for model-based RL without making any assumptions on the structure of the supervisor. We use the state-of-the-art deep MBRL algorithm, PETS, as an improving supervisor and maintain its data efficiency while significantly accelerating policy evaluation. Finally, we evaluate the efficiency of the method by successfully training a policy on a multi-goal reacher task directly on a physical surgical robot. The provided analysis and framework suggests a number of interesting questions regarding the degree to which non-stationary supervisors affect policy learning. In future work, it would be interesting to derive specific convergence guarantees for the converging supervisor setting, consider different notions of supervisor convergence, and study the trade-offs between supervision quality and quantity.

\clearpage
\acknowledgments{This research was performed at the AUTOLAB at UC Berkeley in affiliation with the Berkeley AI Research (BAIR) Lab, Berkeley Deep Drive (BDD), the Real-Time Intelligent Secure Execution (RISE) Lab, and the CITRIS "People and Robots" (CPAR) Initiative.
The authors were supported in part by the Scalable Collaborative Human-Robot Learning (SCHooL) Project, NSF National Robotics Initiative Award 1734633 and by donations from Google, Siemens, Amazon Robotics, Toyota Research Institute, Autodesk, ABB, Samsung, Knapp, Loccioni, Honda, Intel, Comcast, Cisco, Hewlett-Packard and by equipment grants from PhotoNeo, NVidia, and Intuitive Surgical. Any opinions, findings, and conclusions or recommendations expressed in this material are those of the author(s) and do not necessarily reflect the views of the sponsors. We thank our colleagues who provided helpful feedback, code, and suggestions, especially Michael Danielczuk, Anshul Ramachandran, and Ajay Tanwani.}

\begin{small}
\bibliography{corl}  %
\end{small}
\newpage
\appendix
\begin{LARGE}
\begin{center}
\textbf{On-Policy Robot Imitation Learning from a Converging Supervisor Supplementary Material}
\end{center}
\end{LARGE}
\maketitle

\section{Static Regret}

\subsection{Proof of \Cref{thm-static-regret-bound}}
Recall the standard notion of static regret as defined in Definition \autoref{def-static-regret-standard}:
\begin{equation}\mathrm{Regret}_{N}^{S}((\psi_i)_{i=1}^N) = \sum_{i=1}^{N} \left[ l_i(\pi_{\theta_i}, \psi_i) - l_i(\pi_{\theta^\ast}, \psi_i) \right] \text{ where }\theta^\ast = \argmin_{\theta \in \Theta} \sum_{i = 1}^{N} l_i(\pi_{\theta}, \psi_i) \label{eq:std-static}\end{equation}

However, we seek to bound
\begin{equation}\mathrm{Regret}_{N}^{S}(\psi_N) = \sum_{i=1}^{N} \left[ l_i(\pi_{\theta_i}, \psi_N) - l_i(\pi_{\theta^\star}, \psi_N)\right] \text{ where }\theta^\star = \argmin_{\theta \in \Theta} \sum_{i = 1}^{N} l_i(\pi_{\theta}, \psi_N)\label{eq:static-final}\end{equation}
as defined in Definition \autoref{def-static-regret}.

Notice that this corresponds to the static regret of the agent with respect to the losses parameterized by the last observed supervisor $\psi_N$. We can do this as follows:

\begin{align}
    \mathrm{Regret}_{N}^{S}(\psi_N) &= \sum_{i=1}^{N} \left[ l_i(\pi_{\theta_i}, \psi_N) - l_i(\pi_{\theta^\star}, \psi_N)\right] \\&= \sum_{i=1}^{N} \left[ l_i(\pi_{\theta_i}, \psi_N) - l_i(\pi_{\theta^\star}, \psi_N)\right] - \mathrm{Regret}_{N}^{S}((\psi_i)_{i=1}^N) + \mathrm{Regret}_{N}^{S}((\psi_i)_{i=1}^N)\label{t1.1}\\
\begin{split}
&=\sum_{i=1}^{N}\left[ l_i(\pi_{\theta_i}, \psi_N) - l_i(\pi_{\theta_i}, \psi_i)\right]
    +\sum_{i=1}^{N}\left[ l_i(\pi_{\theta^\ast}, \psi_i) - l_i(\pi_{\theta^\star}, \psi_N)\right]\\
    &\quad +\mathrm{Regret}_{N}^{S}((\psi_i)_{i=1}^N)
\end{split}\label{t1.2}\\
\begin{split}
&\leq\sum_{i=1}^{N}\left[ l_i(\pi_{\theta_i}, \psi_N) - l_i(\pi_{\theta_i}, \psi_i)\right]
    +\sum_{i=1}^{N}\left[ l_i(\pi_{\theta^\star}, \psi_i) - l_i(\pi_{\theta^\star}, \psi_N)\right]\\
    &\quad +\mathrm{Regret}_{N}^{S}((\psi_i)_{i=1}^N)
\end{split}\label{t1.3}
\end{align}
Here, inequality \ref{t1.3} follows from the fact that $\sum_{i=1}^{N} l_i(\pi_{\theta^\ast}, \psi_i) \leq \sum_{i=1}^{N} l_i(\pi_{\theta^\star}, \psi_i)$. Now, we can focus on bounding the extra term. Let $h(x, y) = \|x - y\|^2$.
\begin{align}
    &\sum_{i=1}^{N}\left[ l_i(\pi_{\theta_i}, \psi_N) - l_i(\pi_{\theta_i}, \psi_i)\right]
    +\sum_{i=1}^{N}\left[ l_i(\pi_{\theta^\star}, \psi_i) - l_i(\pi_{\theta^\star}, \psi_N)\right]\label{t1.4}\\
    \begin{split}
    &= \sum_{i=1}^{N} \mathbb{E}_{\tau \sim p(\tau|\theta_i)}\left[ \frac{1}{T} \sum_{t=1}^{T} h(\pi_{\theta_i}(s_t^i), \psi_N(s_t^i)) - h(\pi_{\theta_i}(s_t^i), \psi_i(s_t^i)) \right]\\
    & \quad + \sum_{i=1}^{N} \mathbb{E}_{\tau \sim p(\tau|\theta_i)}\left[ \frac{1}{T} \sum_{t=1}^{T} h(\pi_{\theta^\star}(s_t^i), \psi_i(s_t^i)) - h(\pi_{\theta^\star}(s_t^i), \psi_N(s_t^i)) \right]
    \end{split}\label{t1.5}\\
    \begin{split}
    &\leq \sum_{i=1}^{N} \mathbb{E}_{\tau \sim p(\tau|\theta_i)}\left[ \frac{1}{T} \sum_{t=1}^{T} \langle \nabla_\psi h(\pi_{\theta_i}(s_t^i), \psi_N(s_t^i)), \psi_N(s_t^i) - \psi_i(s_t^i)\rangle \right]\\
    &\quad + \sum_{i=1}^{N} \mathbb{E}_{\tau \sim p(\tau|\theta_i)}\left[ \frac{1}{T} \sum_{t=1}^{T} \langle \nabla_\psi h(\pi_{\theta^\star}(s_t^i), \psi_i(s_t^i)), \psi_i(s_t^i) - \psi_N(s_t^i)\rangle \right]
    \end{split}\label{t1.6}
\end{align}
\begin{align}
    \begin{split}
    &= \sum_{i=1}^{N} \mathbb{E}_{\tau \sim p(\tau|\theta_i)}\left[ \frac{1}{T} \sum_{t=1}^{T}\langle 2(\psi_N(s_t^i) - \pi_{\theta_i}(s_t)), \psi_N(s^i_t) - \psi_i(s^i_t) \rangle \right]\\
    &\quad+ \sum_{i=1}^{N} \mathbb{E}_{\tau \sim p(\tau|\theta_i)}\left[ \frac{1}{T} \sum_{t=1}^{T}\langle 2(\psi_i(s_t^i) - \pi_{\theta^\star}(s_t)), \psi_i(s^i_t) - \psi_N(s^i_t) \rangle \right] 
    \end{split}\label{t1.7}\\
    \begin{split}
    &\leq \sum_{i=1}^{N} \mathbb{E}_{\tau \sim p(\tau|\theta_i)}\left[ \frac{1}{T} \sum_{t=1}^{T} 2\|\psi_N(s_t^i) - \pi_{\theta_i}(s_t)\| \|\psi_N(s^i_t) - \psi_i(s^i_t)\| \right]\\ &\quad + \sum_{i=1}^{N} \mathbb{E}_{\tau \sim p(\tau|\theta_i)}\left[ \frac{1}{T} \sum_{t=1}^{T} 2 \|\psi_i(s_t^i) - \pi_{\theta^\star}(s_t)\| \|\psi_i(s^i_t) - \psi_N(s^i_t)\| \right] 
    \end{split}\label{t1.8}\\
    &\leq 4\delta \sum_{i=1}^{N} \mathbb{E}_{\tau \sim p(\tau|\theta_i)}\left[ \frac{1}{T} \sum_{t=1}^{T}\|\psi_N(s^i_t) - \psi_i(s^i_t)\| \right] \label{t1.9}
\end{align}
Equation~\ref{t1.5} follows from applying the definition of the loss function. Inequality~\ref{t1.6} follows from applying convexity of $h$ in $\psi$. Equation~\ref{t1.7} follows from evaluating the corresponding gradients. Inequality~\ref{t1.8} follows from Cauchy-Schwarz and inequality~\ref{t1.9} follows from the action space bound.
Thus, we have:
\begin{align}
    \mathrm{Regret}_{N}^{S}(\psi_N) &\leq 4\delta \sum_{i=1}^{N} \mathbb{E}_{\tau \sim p(\tau|\theta_i)}\left[ \frac{1}{T} \sum_{t=1}^{T}\|\psi_N(s^i_t) - \psi_i(s^i_t)\| \right] +\mathrm{Regret}_{N}^{S}((\psi_i)_{i=1}^N)
\end{align}

\subsection{Proof of \Cref{lemma-static-regret-rate}}
\begin{align}
\forall s \in \mathcal{S},\ \forall N > i, \lVert \psi_i(s) - \psi_N(s)\rVert \leq f_i \text{ where }\lim_{i \rightarrow \infty} f_i = 0 \label{actual_cond}
\end{align}
implies that
\begin{align}
\mathbb{E}_{\tau \sim p(\tau|\theta_i)} \left[ \frac{1}{T} \sum_{t=1}^{T}  \lVert \psi_i(s^i_t) - \psi_N(s^i_t)\rVert \right]\leq f_i \ \forall N > i\in \mathbb{N} \label{suff_cond}
\end{align}

This in turn implies that
\begin{align}
\sum_{i=1}^{N} \mathbb{E}_{\tau \sim p(\tau|\theta_i)} \left[ \frac{1}{T} \sum_{t=1}^{T}  \lVert \psi_i(s^i_t) - \psi_N(s^i_t)\rVert \right]\leq \sum_{i=1}^{N} f_i
\end{align}
\textit{Remark:} For sublinearity, we really only need inequality~\ref{suff_cond} to hold. Due to the dependence of $p(\tau|\theta_i)$ on the parameter $\theta_i$ of the policy at iteration $i$, we tighten this assumption with the stricter Cauchy condition~\ref{actual_cond} to remove the dependence of a component of the regret on the sequence of policies used.

The Additive Ces\`{a}ro's Theorem states that if the sequence $(a_n)_{n=1}^{\infty}$ has a limit, then 

$$\lim_{n \rightarrow \infty} \frac{a_1 + a_2 \hdots a_n}{n} = \lim_{n \rightarrow \infty} a_n$$
Thus, we see that if $\lim_{i \rightarrow \infty} f_i = 0$, then it must be the case that $\lim_{N \rightarrow \infty} \frac{1}{N} \sum_{i = 1}^{N} f_i = 0$. This shows that for some $(f_i)_{i=1}^N$ converging to 0, it must be the case that 
$$\sum_{i=1}^{N} \mathbb{E}_{\tau \sim p(\tau|\theta_i)} \left[ \frac{1}{T} \sum_{t=1}^{T} \lVert \psi_i(s^i_t) - \psi_N(s^i_t)\rVert \right] \leq \sum_{i=1}^{N} f_i = \smallO(N)$$
Thus, based on the regret bound in \Cref{thm-static-regret-bound}, we can achieve sublinear $\mathrm{Regret}_{N}^{S}(\psi_N)$ for any sequence $(f_i)_{i=1}^N$ which converges to 0 given an algorithm that achieves sublinear $\mathrm{Regret}_{N}^{S}((\psi_i)_{i=1}^N)$:

$$\mathrm{Regret}_{N}^{S}(\psi_N) = \mathrm{Regret}_{N}^{S}((\psi_i)_{i=1}^N) + \smallO(N)$$\qed

\section{Dynamic Regret}
\subsection{Proof of \Cref{lemma-dynamic-regret-reduction}}
Recall the standard notion of dynamic regret as defined in Definition \autoref{def-dynamic-regret-standard}:
\begin{equation}\mathrm{Regret}_{N}^{D}((\psi_i)_{i=1}^N) = \sum_{i=1}^{N} \left[ l_i(\pi_{\theta_i}, \psi_i) - l_i(\pi_{\theta_i^\ast}, \psi_i) \right] \text{ where }\theta^\ast_i = \argmin_{\theta \in \Theta} l_{i}(\pi_{\theta}, \psi_i) \label{eq:std-dynamic}\end{equation}

However, we seek to bound
\begin{equation}\mathrm{Regret}_{N}^{D}(\psi_N) = \sum_{i=1}^{N}\left[ l_i(\pi_{\theta_i}, \psi_N) - l_i(\pi_{\theta_i^\star}, \psi_N)\right] \text{ where }\theta^\star_i = \argmin_{\theta \in \Theta} l_{i}(\pi_{\theta}, \psi_N)\label{eq:dynamic-final}\end{equation}
as defined in Definition \autoref{def-dynamic-regret}.

Notice that this corresponds to the dynamic regret of the agent with respect to the losses parameterized by the most recent supervisor $\psi_N$. We can do this as follows:
\begin{align}
    \mathrm{Regret}_{N}^{D}(\psi_N) &= \sum_{i=1}^{N}\left[ l_i(\pi_{\theta_i}, \psi_N) - l_i(\pi_{\theta_i^\star}, \psi_N)\right] \label{l3.1}\\
    \begin{split}
    &= \sum_{i=1}^{N}\left[ l_i(\pi_{\theta_i}, \psi_N) - l_i(\pi_{\theta_i^\star}, \psi_N)\right] - \mathrm{Regret}_{N}^{D}((\psi_i)_{i=1}^N)\\&\quad + \mathrm{Regret}_{N}^{D}((\psi_i)_{i=1}^N)
    \end{split}\label{l3.2}\\
    \begin{split}
&=\sum_{i=1}^{N}\left[ l_i(\pi_{\theta_i}, \psi_N) - l_i(\pi_{\theta_i}, \psi_i)\right]
    +\sum_{i=1}^{N}\left[ l_i(\pi_{\theta_i^\ast}, \psi_i) - l_i(\pi_{\theta_i^\star}, \psi_N)\right]
    \\&\quad +\mathrm{Regret}_{N}^{D}((\psi_i)_{i=1}^N)
    \end{split}\label{l3.3}\\
    \begin{split}
&\leq\sum_{i=1}^{N}\left[ l_i(\pi_{\theta_i}, \psi_N) - l_i(\pi_{\theta_i}, \psi_i)\right]
    +\sum_{i=1}^{N}\left[ l_i(\pi_{\theta_i^\star}, \psi_i) - l_i(\pi_{\theta_i^\star}, \psi_N)\right]
    \\&\quad +\mathrm{Regret}_{N}^{D}((\psi_i)_{i=1}^N)
    \end{split}\label{l3.4}
\end{align}
Here, inequality \ref{l3.4} follows from the fact that $l_i(\pi_{\theta^\ast_i}, \psi_i) \leq l_i(\pi_{\theta^\star_i}, \psi_i)$. Now as before, we can focus on bounding the extra term. Let $h(x, y) = \|x - y\|^2$.
\begin{align}
    &\sum_{i=1}^{N}\left[ l_i(\pi_{\theta_i}, \psi_N) - l_i(\pi_{\theta_i}, \psi_i)\right]
    +\sum_{i=1}^{N}\left[ l_i(\pi_{\theta^\star_i}, \psi_i) - l_i(\pi_{\theta^\star_i}, \psi_N)\right]\label{l1.4}\\
    \begin{split}
    &= \sum_{i=1}^{N} \mathbb{E}_{\tau \sim p(\tau|\theta_i)}\left[ \frac{1}{T} \sum_{t=1}^{T} h(\pi_{\theta_i}(s_t^i), \psi_N(s_t^i)) - h(\pi_{\theta_i}(s_t^i), \psi_i(s_t^i)) \right]\\
    & \quad + \sum_{i=1}^{N} \mathbb{E}_{\tau \sim p(\tau|\theta_i)}\left[ \frac{1}{T} \sum_{t=1}^{T} h(\pi_{\theta^\star_i}(s_t^i), \psi_i(s_t^i)) - h(\pi_{\theta^\star_i}(s_t^i), \psi_N(s_t^i)) \right]
    \end{split}\label{l1.5}\\
    \begin{split}
    &\leq \sum_{i=1}^{N} \mathbb{E}_{\tau \sim p(\tau|\theta_i)}\left[ \frac{1}{T} \sum_{t=1}^{T} \langle \nabla_\psi h(\pi_{\theta_i}(s_t^i), \psi_N(s_t^i)), \psi_N(s_t^i) - \psi_i(s_t^i)\rangle \right]\\
    &\quad + \sum_{i=1}^{N} \mathbb{E}_{\tau \sim p(\tau|\theta_i)}\left[ \frac{1}{T} \sum_{t=1}^{T} \langle \nabla_\psi h(\pi_{\theta^\star_i}(s_t^i), \psi_i(s_t^i)), \psi_i(s_t^i) - \psi_N(s_t^i)\rangle \right]
    \end{split}\label{l1.6}
\end{align}
\begin{align}
    \begin{split}
    &= \sum_{i=1}^{N} \mathbb{E}_{\tau \sim p(\tau|\theta_i)}\left[ \frac{1}{T} \sum_{t=1}^{T}\langle 2(\psi_N(s_t^i) - \pi_{\theta_i}(s_t)), \psi_N(s^i_t) - \psi_i(s^i_t) \rangle \right]\\
    &\quad+ \sum_{i=1}^{N} \mathbb{E}_{\tau \sim p(\tau|\theta_i)}\left[ \frac{1}{T} \sum_{t=1}^{T}\langle 2(\psi_i(s_t^i) - \pi_{\theta^\star_i}(s_t)), \psi_i(s^i_t) - \psi_N(s^i_t) \rangle \right] 
    \end{split}\label{l1.7}\\
    \begin{split}
    &\leq \sum_{i=1}^{N} \mathbb{E}_{\tau \sim p(\tau|\theta_i)}\left[ \frac{1}{T} \sum_{t=1}^{T} 2\|\psi_N(s_t^i) - \pi_{\theta_i}(s_t)\| \|\psi_N(s^i_t) - \psi_i(s^i_t)\| \right]\\ &\quad + \sum_{i=1}^{N} \mathbb{E}_{\tau \sim p(\tau|\theta_i)}\left[ \frac{1}{T} \sum_{t=1}^{T} 2 \|\psi_i(s_t^i) - \pi_{\theta^\star_i}(s_t)\| \|\psi_i(s^i_t) - \psi_N(s^i_t)\| \right] 
    \end{split}\label{l1.8}\\
    &\leq 4\delta \sum_{i=1}^{N} \mathbb{E}_{\tau \sim p(\tau|\theta_i)}\left[ \frac{1}{T} \sum_{t=1}^{T}\|\psi_N(s^i_t) - \psi_i(s^i_t)\| \right] \label{l1.9}
\end{align}
The steps of this proof follow as in the proof of the static regret reduction. Equation~\ref{l1.5} follows from applying the definition of the loss function. Inequality~\ref{l1.6} follows from applying convexity of $h$ in $\psi$. Equation~\ref{l1.7} follows from evaluating the corresponding gradients. Inequality~\ref{l1.8} follows from Cauchy-Schwarz and inequality~\ref{l1.9} follows from the action space bound.
Combining this bound with~\ref{l3.4}, we have our desired result:
\begin{align}
    \mathrm{Regret}_{N}^{D}(\psi_N) &\leq 4\delta \sum_{i=1}^{N} \mathbb{E}_{\tau \sim p(\tau|\theta_i)}\left[ \frac{1}{T} \sum_{t=1}^{T}\|\psi_N(s^i_t) - \psi_i(s^i_t)\| \right] +\mathrm{Regret}_{N}^{D}((\psi_i)_{i=1}^N)
\end{align}\qed

\subsection{Proof of \Cref{lemma-sublinear-standard-dynamic-regret}}
By \Cref{lemma-static-regret-rate}, $$\sum_{i=1}^{N} \mathbb{E}_{\tau \sim p(\tau|\theta_i)} \left[ \frac{1}{T} \sum_{t=1}^{T} \lVert \psi_i(s^i_t) - \psi_N(s^i_t) \rVert \right] = \smallO(N)$$ which implies that 
$$\mathrm{Regret}_{N}^{D}(\psi_N) = \mathrm{Regret}_{N}^{D}((\psi_i)_{i=1}^N) + \smallO(N)$$\qed

\subsection{Predictability of Online Learning Problems}
Next, we establish that the online learning problem defined by the losses defined in Section \ref{sec:framework} is an $(\alpha, \beta)$-predictable online learning problem as defined in \citet{Cheng-COL}. An online learning problem is  $(\alpha, \beta)$-predictable if it satisfies $ \forall \theta \in \Theta$, (1) $l_i(.)$ is $\alpha$ strongly convex in $\theta$, (2) $\lVert \nabla_\theta l_{i+1}(\pi_\theta, \psi_{i+1}) - \nabla_\theta l_{i}(\pi_\theta, \psi_{i}) \rVert \leq \beta \lVert \theta_{i+1} - \theta_i \rVert + \zeta_i$ where $\sum_{i=1}^{N} \zeta_i = \smallO(N)$. Proposition 12 in \citet{Cheng-COL} shows that for $(\alpha, \beta)$-predictable problems, sublinear dynamic regret can be achieved if $\alpha > \beta$. Furthermore, Theorem 3 in \citet{Cheng-COL} shows that if $\alpha$ is sufficiently large and $\beta$ sufficiently small, then sublinear dynamic regret can be achieved by online gradient descent.

\begin{lemma}
\label{lemma-predictability}
If $\forall s \in \mathcal{S},\ \forall N > i, \lVert \psi_i(s) - \psi_N(s) \rVert \leq f_i$ where $\lim_{i \rightarrow \infty} f_i = 0$, the learning problem is $(\alpha, 4G\eta\sup_{a\in\mathcal{A}}\|a\|)$-predictable in $\theta$:  $l_i(\pi_\theta, \psi)$ is $\alpha$-strongly convex by assumption and if Assumption \ref{assum-lipschitz-traj} holds, then $l_i(\pi_\theta, \psi)$ satisfies:
\begin{align*}
&\lVert \nabla_\theta l_{i+1}(\pi_\theta, \psi_{i+1}) - \nabla_\theta l_{i}(\pi_\theta, \psi_{i}) \rVert \leq 4G \eta \sup_{a\in\mathcal{A}}\|a\|\lVert \theta_{i+1} - \theta_i \rVert + \zeta_i \text{ where } \sum_{i=1}^{N} \zeta_i = \smallO(N)
\end{align*}
\end{lemma}

\textbf{Proof of \Cref{lemma-predictability}}
We have bounded $\mathrm{Regret}_{N}^{D}(\psi_N)$ by the sum of $\mathrm{Regret}_{N}^{D}((\psi_i)_{i = 1}^{N})$ and a sublinear term. Now, we analyze $\mathrm{Regret}_{N}^{D}((\psi_i)_{i = 1}^{N})$. We note that we can achieve sublinear $\mathrm{Regret}_{N}^{D}((\psi_i)_{i = 1}^{N})$ if the losses satisfy $$\lVert \nabla_\theta l_{i+1}(\pi_\theta, \psi_{i+1}) - \nabla_\theta l_{i}(\pi_\theta, \psi_{i}) \rVert \leq \beta\lVert\theta_{i+1} - \theta_i\rVert + \zeta_i$$ where $\sum_{i=1}^{N} \zeta_i = \smallO(N)$ by Proposition 12 in \citet{Cheng-COL}.

Note that for $J_\tau(\pi_\theta, \psi) = \frac{1}{T} \sum_{t = 1}^{T} \lVert \psi(s_t) - \pi_\theta(s_t) \rVert^2$, we have
\begin{align}
    \nabla_\theta l_{i}(\pi_\theta, \psi) &= \mathbb{E}_{\tau \sim p(\tau | \theta_i)} \frac{1}{T} \sum_{t = 1}^{T} \nabla_\theta \lVert \psi(s_t) - \pi_\theta(s_t) \rVert^2 \\
    &= \mathbb{E}_{\tau \sim p(\tau | \theta_i)} \nabla_\theta J_\tau(\pi_\theta, \psi)\\
    &=\int p(\tau|\theta_i) \nabla_\theta J_\tau(\pi_\theta, \psi) d \tau\\
    \nabla_\theta J_\tau (\pi_\theta, \psi) &= \frac{1}{T}\sum_{s_t\in \tau} 2\nabla_\theta \pi_\theta(s_t)^T(\pi_\theta(s_t) - \psi(s_t))\\
    &= \frac{2}{T} \nabla_\theta \pi_\theta(\tau)^T (\pi_\theta(\tau) - \psi(\tau))
\end{align}
where 
\begin{align}
    \psi(\tau) = \begin{bmatrix} \psi(s_0) \\ \vdots \\  \psi(s_T) \end{bmatrix},\ 
    \pi_\theta(\tau) = \begin{bmatrix} \pi_\theta(s_0) \\ \vdots \\ \pi_\theta(s_{T}) \end{bmatrix},\ 
    \nabla_\theta\pi_\theta(\tau) = \begin{bmatrix} \nabla_\theta\pi_\theta(s_0) \\ \vdots \\ \nabla_\theta\pi_\theta(s_{T}) \end{bmatrix}
\end{align}
Taking the difference of the above loss gradients, we obtain:
\begin{align}
    &\lVert \nabla_\theta l_{i+1}(\pi_\theta, \psi_{i+1}) - \nabla_\theta l_{i}(\pi_\theta, \psi_{i}) \rVert\\
    &= \bigg\lVert\int p(\tau|\theta_{i+1}) \nabla_\theta J_\tau(\pi_\theta, \psi_{i+1}) d \tau - \int p(\tau|\theta_i) \nabla_\theta J_\tau(\pi_\theta, \psi_{i}) d \tau \bigg\rVert \label{L5.1} \\
    &\leq \int \|p(\tau|\theta_{i+1}) \nabla_\theta J_\tau(\pi_\theta, \psi_{i+1}) - p(\tau|\theta_i) \nabla_\theta J_\tau(\pi_\theta, \psi_{i})\| d \tau \label{L5.2}\\
    \begin{split}
        &= \int \bigg\lVert \frac{2}{T}\nabla_\theta \pi_\theta(\tau)^T(p(\tau|\theta_i) \psi_i(\tau) - p(\tau|\theta_{i+1})\psi_{i+1}(\tau)) \\&\quad + \frac{2}{T}\nabla_\theta \pi_\theta(\tau)^T(p(\tau|\theta_{i+1}) \pi_{\theta}(\tau) - p(\tau|\theta_{i})\pi_{\theta}(\tau)) \bigg\rVert d \tau
    \end{split}\label{L5.3}\\
    \begin{split}
        &\leq \int \bigg\lVert \frac{2}{T}\nabla_\theta \pi_\theta(\tau)^T(p(\tau|\theta_i) \psi_i(\tau) - p(\tau|\theta_{i+1})\psi_{i+1}(\tau)) \bigg\rVert d \tau \\&\quad+ \int \bigg\lVert \frac{2}{T}\nabla_\theta \pi_\theta(\tau)^T\pi_{\theta}(\tau)(p(\tau|\theta_{i+1})  - p(\tau|\theta_{i})) \bigg\rVert d \tau 
    \end{split}\label{L5.3.1}\\
    \begin{split}
        &\leq \int \bigg\lVert \frac{2}{T}\nabla_\theta \pi_\theta(\tau)^T(p(\tau|\theta_i) \psi_i(\tau) - p(\tau|\theta_{i+1})\psi_{i+1}(\tau)) \bigg\rVert d \tau \\&\quad + 2G\sup_{a\in\mathcal{A}}\|a\|\int|p(\tau|\theta_{i+1})  - p(\tau|\theta_{i})| d \tau
    \end{split}\label{L5.3.2}\\
    &\leq \int \bigg\lVert \frac{2}{T}\nabla_\theta \pi_\theta(\tau)^T(p(\tau|\theta_i) \psi_i(\tau) - p(\tau|\theta_{i+1})\psi_{i+1}(\tau)) \bigg\rVert d \tau  + 2G\eta \sup_{a\in\mathcal{A}}\|a\| \|\theta_{i+1} - \theta_i\| \label{L5.3.3}\\
    &\leq \frac{2}{T}G\int\|  p(\tau|\theta_i) \psi_i(\tau) - p(\tau|\theta_{i+1})\psi_{i+1}(\tau)\| d \tau  + 2G\eta \sup_{a\in\mathcal{A}}\|a\| \|\theta_{i+1} - \theta_i\|  \label{L5.4}\\
    \begin{split}
        &= \frac{2}{T}G\int\|  p(\tau|\theta_i) \psi_i(\tau) - p(\tau|\theta_i) \psi_{i+1}(\tau) + p(\tau|\theta_i) \psi_{i+1}(\tau) - p(\tau|\theta_{i+1})\psi_{i+1}(\tau)\| d \tau\\  
        &\quad + 2G\eta \sup_{a\in\mathcal{A}}\|a\| \|\theta_{i+1} - \theta_i\|
    \end{split}\label{L5.5}
\end{align}
\begin{align}
    \begin{split}
        &\leq \frac{2}{T}G\int\|  p(\tau|\theta_i) (\psi_i(\tau) - \psi_{i+1}(\tau))\| + \|(p(\tau|\theta_i)  - p(\tau|\theta_{i+1}))\psi_{i+1}(\tau)\| d \tau\\  
        &\quad+ 2G\eta \sup_{a\in\mathcal{A}}\|a\| \|\theta_{i+1} - \theta_i\|
    \end{split}\label{L5.6}\\
    &\leq \frac{2}{T}G\int p(\tau|\theta_i) \|  \psi_i(\tau) - \psi_{i+1}(\tau)\| d \tau + 4G\eta\sup_{a\in\mathcal{A}}\|a\|\|\theta_{i+1} - \theta_i\| \label{L5.7}\\
    &\leq 2Gf_i\int p(\tau|\theta_i) d \tau + 4G\eta\sup_{a\in\mathcal{A}}\|a\|\|\theta_{i+1} - \theta_i\| \label{L5.8}\\
    &\leq 2Gf_i + 4G\eta\sup_{a\in\mathcal{A}}\|a\|\|\theta_{i+1} - \theta_i\| \label{L5.9}\\
    &= 4G\eta\sup_{a\in\mathcal{A}}\|a\|\|\theta_{i+1} - \theta_i\| + \zeta_i \label{L5.10}
\end{align}
where here $\zeta_i = 2Gf_i$ and we see that $2G \sum_{i=1}^{N} f_i = \smallO(N)$ as desired for some $(f_i)_{i=1}^{N}$ where $\lim_{i \rightarrow \infty} f_i = 0$ as in \Cref{lemma-static-regret-rate}.
Equation~\ref{L5.1} follows from applying definitions. Equation~\ref{L5.2} follows from the triangle inequality. Equation~\ref{L5.3} follows from substitution of the loss gradients. Inequality~\ref{L5.3.1} follows from the triangle inequality and factoring out common terms. Inequality~\ref{L5.3.2} follows from subadditivity, the policy Jacobian and action space bound. Inequality~\ref{L5.3.3} follows from Assumption~\ref{assum-lipschitz-traj}. Equation~\ref{L5.4} follows from subadditivity of the operator norm and the policy Jacobian bound. Equation ~\ref{L5.6} follows from the triangle inequality, and equation~\ref{L5.7} follows from the triangle inequality and Assumption~\ref{assum-lipschitz-traj}. Equations~\ref{L5.8} and ~\ref{L5.10} follow from the convergence assumption of the supervisor and the triangle inequality.\qed

\begin{lemma} \label{lemma-bounded-loss-grads}
Assumption \autoref{assum-bounded-op-norm} implies that the loss function gradients are bounded as follows:
$$\lVert \nabla_{\theta} l_{i}(\pi_{\theta}, \psi) \rVert \leq 2G\delta \ \ \ \forall \theta,\theta_i \in \Theta,\ \forall \psi$$
\end{lemma}
\textbf{Proof of \Cref{lemma-bounded-loss-grads}}
\begin{align*} 
\begin{split}
&\Bigg\lVert \mathbb{E}_{\tau \sim p(\tau|\theta_i)} \left[ \frac{1}{T} \sum_{t=1}^{T} 2(\nabla_{\theta}\pi_{\theta}(s^i_t))^T\left(\pi_{\theta}(s^i_t) - \psi_i(s^i_t)\right) \right] \Bigg\rVert \leq \\
& \mathbb{E}_{\tau \sim p(\tau|\theta_i)} \left[ \frac{1}{T} \sum_{t=1}^{T} \bigg\lVert 2(\nabla_{\theta}\pi_{\theta}(s^i_t))^T\left(\pi_{\theta}(s^i_t) - \psi_i(s^i_t)\right) \bigg\rVert \right]\\
\end{split}
\end{align*}
by convexity of norms $\lVert \cdot \rVert$ and Jensen's inequality.

Then, we have that 
\begin{align*}
\begin{split}
&\lVert (\nabla_{\theta} \pi_{\theta}(s))^T(\pi_{\theta}(s) - \psi(s)) \rVert \leq
\lVert \nabla_{\theta} \pi_{\theta}(s) \rVert \lVert \pi_{\theta}(s) - \psi(s) \rVert \leq G\delta \ \ \forall \theta \in \Theta, \ \forall s \in \mathcal{S}, \ \forall \psi
\end{split}
\end{align*}
due to subadditivity and the assumption that the action space diameter is bounded. Thus, we have that $$\forall \theta,\theta_i \in \Theta,\forall \psi,
\lVert \nabla_{\theta} l_{i}(\pi_{\theta}, \psi) \rVert \leq 2G\delta \qed$$
\subsection{Proof of \Cref{lemma-dynamic-regret-standard}}
From \Cref{lemma-predictability}, the loss gradients are bounded by the sum of a Lipschitz-type term and a sublinear term, satisfying the conditions for Proposition 12 from \citet{Cheng-COL}. Thus, by Proposition 12 from \citet{Cheng-COL}, we see that as long as $\alpha > 4G\eta \sup_{a \in \mathcal{A}} \lVert a \rVert$, there exists an algorithm that can achieve sublinear $\mathrm{Regret}_{N}^{D}((\psi_i)_{i=1}^N)$. An example of an algorithm that achieves sublinear dynamic regret under this condition is the greedy algorithm \cite{Cheng-COL}: $\theta_{i+1} = \argmin_{\theta \in \Theta} l_i(\pi_\theta, \psi_i)$.

Define $\beta = 4G\eta \sup_{a \in \mathcal{A}} \lVert a \rVert$, $\lambda = \beta / \alpha$, and $\xi_i = \zeta_i / \alpha$. For the greedy algorithm, the result can be shown in a similar fashion to Theorem 3 of \citet{Cheng-COL}:
\begin{align*}
    \|\theta_i^* - \theta_i \| = \| \theta_i^* - \theta_{i - 1}^*\| \leq \lambda \|\theta_{i} - \theta_{i - 1} \| + \frac{\zeta_i}{\alpha}  \leq \lambda ^i\|\theta_1 - \theta_0\| + \sum_{j = 1}^{i} \lambda^{i - j } \xi_{j}
\end{align*}
where the first inequality follows from Proposition 1 of \citet{Lee2019ADR} and the second inequality follows from repeated application of the same proposition. Summing from $1$ to $N$ with $\zeta_i = 2Gf_i$ as in the proof of \Cref{lemma-dynamic-regret-standard}, we have
\begin{align*}
    \sum_{i = 1}^N \sum_{j = 1}^{i} \lambda ^{i - j } \xi_{j} & \leq  \sum_{i = 1}^N \xi_i (1 + \lambda + \lambda^2 + \ldots) \leq   \frac{1}{1 - \lambda} \sum_{i = 1}^N \xi_i = \frac{2G}{\alpha(1 - \lambda)} \sum_{i = 1}^N f_i
\end{align*}
Thus, if $\sum_{i = 1}^N f_i = \smallO(N)$, we can show that the greedy algorithm achieves sublinear $\mathrm{Regret}_{N}^{D}((\psi_i)_{i=1}^N)$ by using the Lipschitz continuity of the losses as shown in the proof of \Cref{lemma-bounded-loss-grads} if the parameter space diameter is bounded as follows: $D = \sup_{\theta, \theta'\in\Theta} \|\theta - \theta'\|$.
\begin{align*}
    \mathrm{Regret}_{N}^{D}((\psi_i)_{i=1}^N) & \leq 2G\delta \sum_{i = 1}^N \| \theta_i - \theta_i^*\| \\
    & \leq 2G\delta \left(D\sum_{i = 1}^N \lambda^i +  \frac{2G}{\alpha(1 - \lambda)} \sum_{i = 1}^N f_i \right) \\
    & \leq 2G\delta \left(\frac{D}{1 - \lambda} +  \frac{2G}{\alpha(1 - \lambda)} \sum_{i = 1}^N f_i\right)  \\
    & = \smallO(N)
\end{align*}

For the last part of the lemma, the fact that online gradient descent achieves sublinear $\mathrm{Regret}_{N}^{D}((\psi_i)_{i=1}^N)$ follows directly from applying Theorem 3 from \citet{Cheng-COL} with $\frac{4G\eta\sup_{a \in \mathcal{A}}\|a\|}{\alpha} > \frac{\alpha}{2\gamma}$ if the losses are $\gamma$-smooth in $\theta$. \qed

\subsection{Proof of \Cref{thm-dynamic-regret-bound}}
The proof follows immediately from combining the result of \Cref{lemma-sublinear-standard-dynamic-regret} and \Cref{lemma-dynamic-regret-standard}. \qed

\section{Training Details}
\subsection{CSF Learner}
For the linear policy, the CSF learner is trained via linear regression with regularization parameter $\alpha=1$. For the neural network policy, the CSF learner is represented with an ensemble of 5 neural networks, each with 2 layers with 20 hidden units and swish activations.
\subsection{PETS}
PETS learns an ensemble of neural network dynamics models using sampled transitions and updates them on-policy to better reflect the dynamics local to the learned policy's state distribution. We use the implementation from \cite{PETS_github}. MPC is run over the learned dynamics to select actions for the next iteration. For all environments, a probabilistic ensemble of 5 neural networks with 3 hidden layers, each with 200 hidden units and swish activations are used to represent the dynamics model. The TS-$\infty$ sampling procedure is used for planning. We use an MPC planning horizon of length 25 for all environments and 1 initial random rollout to seed the learned dynamics model. \citet{handful-of-trials} contains further details on training PETS.
\subsection{SAC}
We use the rlkit implementation \cite{rlkit_github} of soft actor critic with the following parameters: batch size = $128$, discount factor = $0.99$, soft target $\tau$ = $0.001$, policy learning rate = $0.0003$, Q function learning rate = $0.0003$, value function learning rate = $0.0003$, and replay buffer size = $1000000$. All networks are two-layer multi-layer perceptrons with 300 hidden units. 
\subsection{TD3}
We use the rlkit implementation \cite{rlkit_github} of TD3 with the following parameters: batch size = $128$, discount factor = $0.99$, and replay buffer size = $1000000$. The exploration strategy consists of adding Gaussian noise $\mathcal{N}(0,\,0.1)$ to actions chosen by the policy. All networks are two-layer multi-layer perceptrons with 300 hidden units.
\subsection{ME-TRPO}
We model both the policy and dynamics with neural networks, using an ensemble of dynamics models to avoid exploitation of model bias. We use the ME-TRPO implementation from \cite{ME-TRPO_github} with the following hyperparameters: batch size=128, discount factor=1, and learning rate =.001 for both the policy and dynamics. The policy network has two hidden layers with 64 units each and all dynamics networks have two hidden layers with 512 units each and ReLU activation. 

\section{Experimental Details}
\subsection{Simulated Experiments}
Both simulated experiments involve manipulation tasks on a simulated PR2 robot and are from the provided code in \citet{handful-of-trials}. Both are implemented as 7-DOF torque control tasks. For all tasks, we plot the sum of rewards for each training episode.
\subsection{Physical Experiments}
Both physical experiments involve delta-position control in 3D space on the daVinci surgical system, which is cable driven and hard to precisely control, making it difficult to reliably reach a desired pose without appropriate compensation \cite{cable-driven}. The CSF learner policy and supervisor dynamics are modeled by 3 hidden-layer feed-forward neural networks with 200 hidden units each. The tasks involve guiding the end effectors to targets in the workspace and isotropic concave quadratic rewards are used. For all tasks, we plot the sum of rewards for each training episode. For multi-arm experiments, the arms are limited to subsets of the state space where collisions are not possible. We are investigating modeling arm collisions for future work. Since the da Vinci surgical system has relatively limited control frequency, although the CSF learner often enables significantly faster query time than PETS, the improvement in policy evaluation time was somewhat less significant due to physical hardware constraints. In future work, we plan to implement the proposed algorithm on a robot with higher frequency control capability.

\end{document}